\documentclass{article}

\usepackage{microtype}
\usepackage{graphicx}
\usepackage{subcaption}
\usepackage{booktabs} % for professional tables

\usepackage{hyperref}

 \usepackage[accepted]{icml2026}

\usepackage{amsmath}
\usepackage{amssymb}
\usepackage{mathtools}
\usepackage{amsthm}

\usepackage{subcaption}
\usepackage{comment}

\usepackage{enumitem}
\usepackage[capitalize,noabbrev]{cleveref}

\theoremstyle{plain}
\newtheorem{theorem}{Theorem}[section]

\theoremstyle{definition}

\theoremstyle{remark}

\def\beq{\begin{equation}}
\def\eeq{\end{equation}}
\def\beqn{\begin{eqnarray}}
\def\eeqn{\end{eqnarray}}
\def\beqns{\begin{eqnarray*}}
\def\eeqns{\end{eqnarray*}}
\def\argmax{\mathop{\rm argmax}}
\def\argmin{\mathop{\rm argmin}}

\def\EE{{\mathbb{E}}}

\def\SD{{\cal D}}

\def\SP{{\cal P}}

\def\SX{{\cal X}}
\def\SY{{\cal Y}}

\def\SN{{\cal N}}

\def\SD{{\cal D}}

\renewcommand{\Re}{\mathbb{R}}
\def\veps{\varepsilon}

\def\equ#1{(\ref{#1})}

\def\one#1{\mathbf{1}{\{ {#1}\}}}
\def\entropy#1{{\mathbb{H}\left ( {#1} \right ) }}
\def\kl#1{{{\rm D}_{\rm KL}\left ( {#1}\right )}}

\def\##1{\relax\ifmmode\mathchoice      % italic bold vector, e.g. $\#{x}$
{\mbox{\boldmath$\displaystyle#1$}}
 {\mbox{\boldmath$\textstyle#1$}}
{\mbox{\boldmath$\scriptstyle#1$}}
{\mbox{\boldmath$\scriptscriptstyle#1$}}\else
\hbox{\boldmath$\textstyle#1$}\fi}
\newtheorem{example}{Example}

\usepackage[textsize=tiny]{todonotes}

\icmltitlerunning{Epistemic Reject Option Prediction}

\begin{document}

\twocolumn[
  \icmltitle{Epistemic Reject Option Prediction}



  \icmlsetsymbol{equal}{*}

  \begin{icmlauthorlist}
    \icmlauthor{Vojtech Franc}{equal,ctu}
    \icmlauthor{Jakub Paplham}{equal,ctu}
  \end{icmlauthorlist}

  \icmlaffiliation{ctu}{Department of Cybernetics, FEE Czech Technical University in Prague}

  \icmlcorrespondingauthor{Vojtech Franc}{xfrancv@fel.cvut.cz}
  %\icmlcorrespondingauthor{Firstname2 Lastname2}{first2.last2@www.uk}

  \icmlkeywords{}

  \vskip 0.3in
]



\printAffiliationsAndNotice{}  % no special notice (required even if empty)

\begin{abstract}
In high-stakes applications, predictive models must not only produce accurate predictions but also quantify and communicate their uncertainty. Reject-option prediction addresses this by allowing the model to abstain when prediction uncertainty is high. Traditional reject-option approaches focus solely on aleatoric uncertainty, an assumption valid only when large training data makes the epistemic uncertainty negligible. However, in many practical scenarios, limited data makes this assumption unrealistic. This paper introduces the epistemic reject-option predictor, which abstains in regions of high epistemic uncertainty caused by insufficient data. Building on Bayesian learning, we redefine the optimal predictor as the one that minimizes expected regret -- the performance gap between the learned model and the Bayes-optimal predictor with full knowledge of the data distribution. The model abstains when the regret for a given input exceeds a specified rejection cost. To our knowledge, this is the first principled framework that enables learning predictors capable of identifying inputs for which the available training data is insufficient to support well-informed predictions.
\end{abstract}

\section{Introduction}

Predictive models learned from data lie at the core of most practical AI systems. These models are typically trained to minimize the average prediction error. However, in high-stakes applications such as medicine, it is crucial not only to achieve low prediction error but also to quantify the uncertainty of individual predictions and effectively communicate this uncertainty to the end user.

Prediction uncertainty arises from two primary sources: aleatoric uncertainty and epistemic uncertainty~\cite{Hullermeier-Intro-ML2021}. Aleatoric uncertainty stems from the inherent randomness in the relationship between the inputs and the target outputs. It reflects noise in the data and is considered irreducible. In contrast, epistemic uncertainty arises due to the model being trained on a finite dataset instead of having access to the true data-generating distribution. Unlike aleatoric uncertainty, epistemic uncertainty is reducible with more data.

\vspace{-0.1cm}

Prediction uncertainty can be leveraged to support decision-making in various ways. A widely used approach is reject-option prediction, where the model abstains from predicting when uncertainty is high, thus indicating an elevated risk of an incorrect outcome. Traditional reject-option predictors~\cite{Chow-RejectOpt-TIT1970} consider only aleatoric uncertainty, a setting we term aleatoric reject-option prediction. Existing methods for training these predictors often rely on empirical risk minimization or maximum likelihood estimation, assuming a training dataset large enough to disregard epistemic uncertainty~\cite{Hendrickx-MLwithRejectOpt-ML2024}.

\vspace{-0.1cm}

In practical applications, training data is inherently finite, and consequently, ignoring epistemic uncertainty is often unjustified. Bayesian learning provides a principled methodology for naturally incorporating both aleatoric and epistemic uncertainties~\cite{Bishop-PRML-2006,Gelman-Bayesian-2013}. Its application to reject-option prediction enables the development of models we call Bayesian reject-option predictors. Such models are designed to abstain from making a prediction when the overall predictive uncertainty, encompassing both its aleatoric and epistemic sources, is deemed too high.

\vspace{-0.1cm}

In this paper, we introduce a novel framework for constructing epistemic reject-option predictors that abstain when the learned predictor is expected to perform significantly worse than the Bayes-optimal predictor. We formalize this notion by extending the standard Bayesian learning framework to minimize expected regret rather than expected loss. Expected regret, which we use as a measure of epistemic uncertainty, is defined as the performance gap between the learned predictor and the Bayes-optimal predictor with full knowledge of the data-generating distribution. The resulting predictor abstains whenever this regret for a given input exceeds a user-specified rejection cost.

%In this paper, we introduce a novel framework for constructing epistemic reject-option predictors that abstain when the learned predictor is expected to perform significantly worse than the Bayes-optimal predictor. To formalize this notion, we extend the standard Bayesian learning framework by replacing the conventional objective of minimizing expected loss with the minimization of expected regret. The regret, which we use as a measure of epistemic uncertainty, is defined as the performance gap between the learned predictor and the Bayes-optimal predictor, which has full knowledge of the true data-generating distribution. The resulting predictor abstains whenever the regret for a given input exceeds a user-specified rejection cost.

To illustrate the distinction between existing reject-option strategies and the proposed approach, we consider the following example.

\begin{example}
Imagine a system trained to predict house prices from their location. The training data comes primarily from the city center, which has many past sales, so the model is well informed in this region. However, prices in the center vary widely due to unobserved factors, representing aleatoric uncertainty.
\vspace{-0.08cm}

An \textit{aleatoric reject-option} predictor abstains in \textit{locations where prices are highly variable} and predicts where prices are stable. It is reliable only for locations drawn from the city center that are well covered by the training data.
\vspace{-0.08cm}

A \textit{Bayesian reject-option} predictor rejects predictions both in \textit{unfamiliar suburbs}, due to limited training data, and in \textit{well-sampled neighborhoods where prices are volatile}.

\vspace{-0.08cm}

The proposed \textit{epistemic reject-option} predictor \textit{abstains in unfamiliar suburbs}, explicitly signaling that the available training data is insufficient. It continues to predict in the city center despite high price volatility; since it has sufficient data to claim that no predictor can perform substantially better there.
\end{example}

\vspace{-0.05cm}

The main contributions of this work are:
\vspace{-0.3cm}
\begin{enumerate}[leftmargin=1.1em,itemsep=1pt]
\item We propose a principled framework for constructing reject-option predictors whose accept–reject decisions are based solely on epistemic uncertainty. We formulate epistemic reject-option prediction as a decision-making problem with a clear and interpretable objective. To our knowledge, this is the first framework that systematically enables predictors to identify inputs for which the available training data is insufficient to support well-informed predictions.
\item We provide a decision-theoretic justification for entropy-based and variance-based measures of epistemic uncertainty commonly used in Bayesian neural networks~\cite{Depeweg-Decompos-ICML2018}. Specifically, we show that these widely used uncertainty measures correspond to conditional regret—instantiated for different loss functions—which governs the accept–reject decisions of the theoretically optimal epistemic reject-option predictor.
\item We empirically evaluate the proposed approach alongside existing reject-option predictors in a controlled setting where data are generated from known distributions, allowing the optimal predictors to be constructed and fairly evaluated. The experimental results support our theoretical findings.

\vspace{-0.2cm}

\end{enumerate}

\section{Preliminaries}

In this section, we review existing methods for constructing reject-option predictors, illustrating them with a simple running example. 
%We cover standard approaches, including maximum likelihood and Bayesian learning, and show how they apply to reject-option prediction. 
While the material is well established (e.g.,~\cite{Bishop-PRML-2006}), we present it using consistent notation and terminology to provide the necessary context for our novel framework of epistemic reject-option predictors.%, which builds upon these foundations.

\subsection{Aleatoric Reject-Option Prediction}
\label{sec:AleatorRejectOption}

Let $\mathcal{X}$ denote the input space and $\mathcal{Y}$ the target space. Assume that input-target pairs $(x,y)$ are i.i.d. samples from a distribution $p(x,y)$ defined over $\mathcal{X}\times\mathcal{Y}$.
The classical reject-option prediction framework~\cite{Chow-RejectOpt-TIT1970} aims to construct a predictor
$q\colon \mathcal{X}\to \mathcal{Y}\cup\{\text{reject}\}$
that for given $x\in\SX$ either outputs a prediction of the target $q(x)\in\mathcal{Y}$ or abstains from predicting, $q(x)=\text{reject}$, on inputs $x\in\mathcal{X}$ that are likely to be predicted incorrectly.
We represent a reject-option predictor $q$ by a pair of functions $(h,c)$, where
$h\colon \mathcal{X}\to\mathcal{Y}$ is a base predictor and
$c\colon \mathcal{X}\to\{0,1\}$ is a selector, such that
\[
   q(x) = (h,c)(x) =
   \begin{cases}
     h(x), & \text{if }\;\; c(x)=1 \:,\\
     \text{reject}, & \text{if }\;\; c(x)=0 \:.
   \end{cases}
\]
Let $\ell\colon\SY\times\SY\rightarrow\Re_+$ be a prediction loss and $\veps\in\Re_+$ a rejection cost. Then, the decision of a reject-option predictor $q=(h,c)$ on a sample $x\sim p(x)$ is penalized as follows:
%\noindent 1. 
\begin{enumerate}
    \item If it predicts, $c(x)=1$, the penalty equals the conditional risk incurred by predicting $h(x)$ at $x$:
  \[
    a(x) = \EE_{y\sim p(y|x)}\big [ \ell(y,h(x)) \big ]\:,
  \]
%\noindent 2. 
\item If it rejects, $c(x)=0$, the penalty equals $\veps$.
\end{enumerate}

The overall performance of the reject-option predictor $q=(h,c)$ is then evaluated by:
\[
   r_{\varepsilon}(h,c) =
   \mathbb{E}_{x\sim p(x)}
   \Big[
      c(x)\, a(x) + \neg c(x)\,\varepsilon
   \Big] \:,
\]
where we use shortcut $\neg c(x) = 1 - c(x)$. The optimal aleatoric reject-option predictor $q_a=(h_*,c_a)$ that minimizes $r_{\varepsilon}(h,c)$ consists of the Bayes-optimal predictor 
\begin{equation}
  \label{equ:BayesPredictor}
  h_*(x) = \argmin_{\hat{y}\in\SY}\EE_{y\sim p(y\mid x)}\big [ \ell(y,\hat{y})\big ]
\end{equation}
and the selector $c_a(x) =\one{a_*(x) \leq \varepsilon}$ where
\begin{equation}
  \label{equ:TrueAleatorUncertainty}
    a_*(x) = \mathbb{E}_{y\sim p(y\mid x)} \big[ \ell(y,h_*(x) ) \big]\:.
\end{equation}
The conditional risk $a_*(x)$ serves here as a measure of the aleatoric uncertainty of the Bayes-optimal predictor $h_*(x)$ with full knowledge of the data distribution $p(x,y)$. We refer to $a_*(x)$ as the true aleatoric uncertainty.

\begin{figure}[ht!]
    \centering
    % --- Left Column (contains the single figure on the left) ---
    % The [c] option vertically centers this minipage relative to the right one.
    \begin{minipage}[l]{\linewidth}
        \centering
        % This is FIGURE 1
        \includegraphics[width=\linewidth]{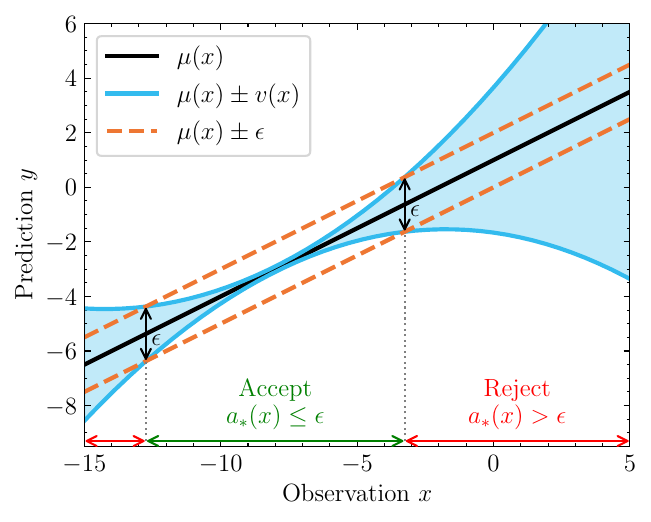}
        \captionof{figure}{The aleatoric reject-option predictor outputs the prediction $h_*(x)$ in regions where the conditional risk $a_*(x)$ does not exceed the rejection cost $\varepsilon$, and abstains otherwise. This example considers the squared loss $\ell(y, \hat{y}) = (y - \hat{y})^2$, the reject cost $\veps=1$ and full knowledge of data distribution $p(x, y)$ defined in Example~\ref{example:1}. Under this loss, the Bayes optimal predictor $h_*(x)$ returns the conditional mean $\mu(x)$, and the aleatoric uncertainty $a_*(x)$ equals the conditional variance $v(x)$.
        }
        \label{fig:aleatoricRejectOption}
    \end{minipage}
\end{figure}

\begin{example}
\label{example:1}
Let $\mathcal{X} = \mathbb{R}$ and $\mathcal{Y}=\Re$. Assume $(x, y) \in \mathcal{X} \times \mathcal{Y}$ is a sample from $p(x, y) = p(x)\, p(y \mid x)$, where
$p(x) = \mathcal{N}(x \mid 0,\, 1)$ is a standard normal distribution and $p(y \mid x) = \mathcal{N}(y \mid \mu(x),\ v(x))$ is normal with input-dependent mean $\mu(x)=0.5x + 1$ and variance $v(x) = 0.1 + 0.04(x + 8)^2$. 
Assume the squared loss $\ell(y, \hat{y}) = (y - \hat{y})^2$. Then, the Bayes-optimal predictor~\equ{equ:BayesPredictor} equals the conditional mean:
$$
h_*(x) = \mathbb{E}_{y \sim p(y \mid x)}[y] = \mu(x)=0.5x + 1.
$$
Accordingly, the true aleatoric uncertainty~\equ{equ:TrueAleatorUncertainty} is equal to the conditional variance of $y$ given $x$:
$$
a_*(x) = \mathrm{Var}_{y \sim p(y \mid x)}[y] = v(x) = 0.1 + 0.04(x + 8)^2.
$$
Figure~\ref{fig:aleatoricRejectOption} illustrates the optimal aleatoric reject-option $q_a=(h_*,c_a)$ using $h_*(x)$ and $a_*(x)$ from this example. %Note that the resulting predictor is independent of the marginal distribution $p(x)$; however, $p(x)$ influences learned predictors, as demonstrated in following examples.
\end{example}

\begin{figure*}[ht!]
    \begin{subfigure}[b]{0.47\linewidth}        
      \includegraphics[width=\linewidth]{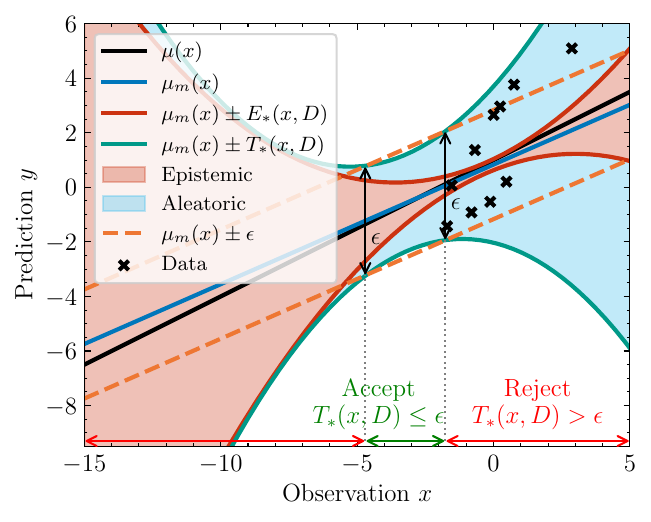}
      \caption{Bayesian reject-option predictor}
      \label{fig:BayesianRejectOpt}
    \end{subfigure}
    \hfill
    \begin{subfigure}[b]{0.47\linewidth}
       \includegraphics[width=\linewidth]{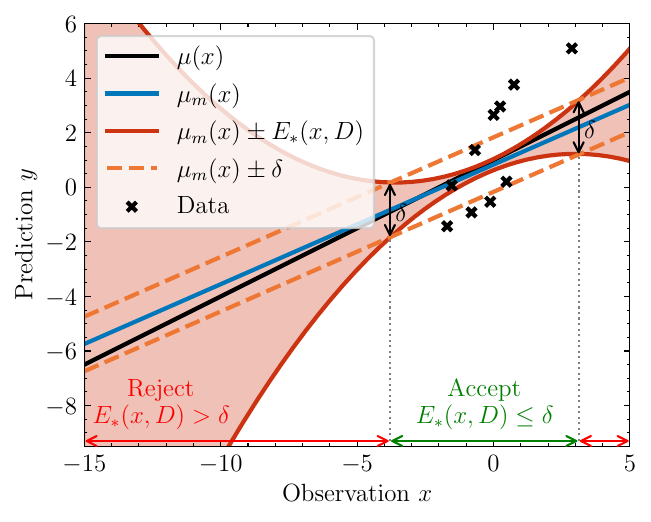}
       \caption{Epistemic reject-option predictor}
      \label{fig:EpistemicRejectOpt}
    \end{subfigure}
    \caption{Fig~\ref{fig:BayesianRejectOpt} illustrates the Bayesian reject-option predictor $Q_B=(H_*,C_T)$, which outputs the prediction $H_*(x, D)$ in regions where the total uncertainty $T_*(x, D)$ does not exceed the rejection cost $\varepsilon = 2$, and abstains otherwise. Fig~\ref{fig:EpistemicRejectOpt} shows the epistemic reject-option predictor $Q_E=(H_*,C_E)$, which outputs the prediction $H_*(x,D)$ in regions where the epistemic uncertainty $E_*(x,D)$ does not exceed the rejection cost $\delta=1$, and abstains elsewhere. This example considers the setting of Example~\ref{example:BayesianRejectOption}, with training set $D$ consisting of $m = 10$ examples drawn from $p(x,y)$ specified in Example~\ref{example:1}. Under this setup, the predictor is given by $H_*(x, D) = \#\phi(x)^T \#\mu_m$, where $\#\mu_m$ denotes the mean of the predictive distribution, serving as an estimate of the true mean $\mu(x)$. The total uncertainty is computed as $T_*(x, D) = \#\phi(x)^T \#\Sigma_m \#\phi(x) + v(x)$, which decomposes into the sum of aleatoric and epistemic uncertainties: $T_*(x, D) = A_*(x, D) + E_*(x, D)$, where $A_*(x, D) = v(x)$ and $E_*(x, D) = \#\phi(x)^T \#\Sigma_m \#\phi(x)$. 
    }
    \label{fig:BayesEpistemicRejOpt}
\end{figure*}

\subsection{Maximum-Likelihood learning}
\label{sec:MLLearning}

The aleatoric reject-option framework (c.f. Sec~\ref{sec:AleatorRejectOption}) assumes that the true data distribution $p(x,y)$ is known, which is usually not the case. Instead, we are often given a training set $D=((x_i,y_i)\in\SX\times\SY\mid i=1,\ldots,m)$ assumed to be i.i.d. drawn from $p(x,y)$. 

In this setup, a common approach to approximate the optimal aleatoric reject-option predictor $q_a=(h_*,c_a)$ is based on the Maximum-Likelihood (ML) learning. It requires to define parametric family of distributions $\{p(x,y\mid \theta )\mid \theta \in \Theta \}$ assumed to contain the data distribution $p(x,y)$. We further use an assumption common in supervised learning that the marginal distribution $p(x)$ is not parametric, i.e. $p(x,y\mid \theta)=p(x)\,p(y\mid x,\theta)$.

Given the training set $D$, the ML learning outputs parameter estimate $\hat{\theta}\in\Theta$ by maximizing the log-likelihood:
\[
  l(\theta) = \log p(D\mid \theta)\quad\text{where}\quad p(D\mid \theta) = \prod_{i=1}^m p(x_i,y_i\mid \theta) \:. 
\]
For example, training classification neural networks with soft-max output layer and cross-entropy loss can be interpreted as special instances of ML learning. The output layer of the trained NN then represents $p(y\mid x,\hat{\theta})$. 

The optimal aleatoric reject-option predictor $(h_*,c_a)$ is then approximated by $(\hat{h},\hat{c})$ where $\hat{h}=h(x,\hat{\theta})$ and $\hat{c}=c(x,\hat{\theta})$ are the plug-in optimal predictor and selector, respectively:
\begin{eqnarray}
   \label{equ:BayesOptimalGivenTheta}
    h(x,\theta) & = &\argmin_{\hat{y}\in\SY} \EE_{y\sim p(y\mid x,\theta)}[\ell(y,\hat{y})] \,,\\
    c(x,\theta) &= &\one{a(x,\theta) \leq \veps}\:.\\
    a(x,\theta) &= &\min_{\hat{y}\in\SY} \EE_{y\sim p(y\mid x,\theta)}[\ell(y,\hat{y})] \,.
\end{eqnarray}

ML learning yields a point estimate $(\hat{h},\hat{c})$ inferred from the training data $D$. This approach accounts only for aleatoric uncertainty which is justifiable when the epistemic uncertainty is negligible. This assumption is reasonable when the training set size $m$ is sufficiently large, as the ML estimate is under some regularity conditions statistically consistent (e.g.~\cite{Wackerly-2002}), i.e. $\hat{\theta}$ converges in probability to $\theta_*$ defining the true distribution $p(x,y\mid \theta_*)=p(x,y)$. 

\begin{example}Assume $p(x,y|\#\theta)=p(x)\,p(y|x,\#\theta)$
and $p(y|x,\#\theta)=\SN\big (y|x\theta_1+\theta_0,v(x)\big )$, where $\#\theta=(\theta_0,\theta_1)\in\Theta=\Re^2$ are unknown parameters to be learned and $v(x)=0.1 + 0.04(x + 8)^2$ is a known variance.  Given i.i.d. training set $D=((x_i,y_i)\in\Re^2|i=1,\ldots,m)$, the ML estimate of the parameters reads:
\[
   \hat{\#\theta}= \big (\#X^T\#\Sigma^{-1}\#X\big )^{-1} \#X\#\Sigma^{-1}\#y \:,
\]
where 
\begin{equation}
\label{equ:DataXy}
\#X=\left [ \begin{array}{cc}x_1& 1\\ \vdots&\vdots \\ x_m&1\\ \end{array}\right]\in\Re^{m\times 2}\,, \quad\#y=\left [ \begin{array}{c} y_1\\ \vdots \\ y_m \end{array}\right ]\in\Re^{m}\,, 
\end{equation}
represent the training data and 
\begin{equation}\label{equ:DataNoise}
\#\Sigma = {\rm diag}\big( [v(x_1),\ldots,v(x_m)]\big )\in\Re_{+}^{m\times m}
\end{equation}
the known heteroscedastic noise at the training inputs.

\end{example}

\subsection{Bayesian Learning}

Bayesian learning (BL)~\cite{Bishop-PRML-2006,Gelman-Bayesian-2013} is an established framework to infer predictors from data while modeling both the aleatoric and the epistemic uncertainty. It relies on a parametric generative model in which the model parameters are treated as random variables. Let $\theta \in \Theta$ denote the parameter, $D \in \SD = (\SX \times \SY)^m$ the training data, $x \in \SX$ an input observation, and $y \in \SY$ its associated target. The tuple $(\theta,D,x,y)$ is considered a sample from a joint distribution $p(\theta,D,x,y)$ defined over $\Theta\times\SD\times\SX\times\SY$. 

The goal of BL is to construct a predictor $H \colon \SX \times (\SX \times \SY)^m \rightarrow \SY$ that, given data $(x,D)$ composed of a training set $D$ and a new input $x$, outputs a prediction $H(x, D) \in \SY$. The performance of such a predictor is evaluated by:
\begin{multline}
\label{equ:BayesLearnExpRisk}
R(H) = 
 \EE_{(\theta,D,x,y)\sim p(\theta,D,x,y)} \big[\ell(y, H(x, D))\big]\;.
\end{multline}
The optimal Bayesian predictor which minimizes $R(H)$ is:
\begin{equation}
\label{equ:BLOptBayes}
H_*(x, D) = \argmin_{\hat{y} \in \SY} \EE_{y\sim p(y|x,D)}\big [ \ell(y, \hat{y}) \big ] \:,
\end{equation}
where $p(y\mid x,D)$ is called the predictive distribution.

\begin{example}
\label{example:BLregression}Assume $p(x,y|\#\theta)=p(x)\,p(y|x,\#\theta)$
and $p(y|x,\#\theta)=\SN(y|x\theta_1+\theta_0,v(x))$, with unknown parameters $\#\theta=(\theta_0,\theta_1)\in\Re^2$, and known variance $v(x)=0.1+0.04(x+8)^2$. Let prior distribution over the parameters be 2D normal distribution $p(\#\theta)=\SN(\#\theta|\#0,\#\Lambda)$, where $\#\Lambda={\rm diag}([\lambda_1,\lambda_2])$ is a diagonal covariance matrix. Let the training set be $D=((x_i,y_i)\in\Re^2)|i=1,\ldots,m)$, drawn i.i.d.  
Under this setup, the predictive distribution is normal $p(y|x,D)=\SN\big(y|\#\phi(x)^T\#\mu_m,\#\phi(x)^T\#\Sigma_m\#\phi(x)+v(x)\big)$, where $\#\phi(x)=[x,1]^T$, and the posterior parameters are given by
\[
\#\Sigma_m=\big(\#X^T\#\Sigma^{-1}\#X+\#\Lambda^{-1} )^{-1}\,,\quad \#\mu_m=\#\Sigma_m\#X^T\#\Sigma^{-1}\#y \:.
\]
Here, $\#X$, $\#y$ and $\#\Sigma$ are defined in equations~\equ{equ:DataXy} and~\equ{equ:DataNoise}.
%, and $\#\Lambda={\rm diag([\lambda_1^{-2},\lambda_2^{-2}])}$. 
Assuming the squared loss $\ell(y,\hat{y})=(y-\hat{y})^2$, the optimal Bayesian predictor corresponds to the mean of the predictive distribution $p(y|x,D)$:
\begin{equation}
\label{equ:BLpredLinReg}
   H_*(x,D) = \EE_{y\sim p(y|x,D)}[y]=\#\phi(x)^T\#\mu_m \:.
\end{equation}
\end{example}

\subsection{Bayesian Reject-Option Prediction}

The predictive distribution inferred by BL can be readily used to construct the reject-option predictor. Specifically, let us consider reject-option predictor $Q\colon\SX\times(\SX\times\SY)^m\rightarrow\SY\cup \{\rm reject\}$, that we parametrize by a pair of functions $(H,C)$, where $H\colon
\SX \times (\SX \times \SY)^m \rightarrow \SY$ is a predictor and $C\colon\SX \times (\SX \times \SY)^m \rightarrow \{0,1\}$ a selector defining $Q$ via
\begin{equation}
  \label{equ:BLRejectOptionPred}
   Q(x,D) = \left \{ \begin{array}{rcl}
      H(x,D) & \text{if} & C(x,D) = 1 \,,\\
      \text{reject} & \text{if} & C(x,D) = 0\:.
   \end{array}
   \right .
\end{equation}
Let $\ell\colon\SY\times\SY\rightarrow\Re_+$ be a prediction loss and  $\veps\in\Re_+$ a rejection cost. Then, the output of the Bayesian reject-option predictor $Q=(H,C)$ on a data sample $(x,D)\sim p(x,D)$ is penalized as follows:

\noindent 1. If it predicts, $C(x,D)=1$, the penalty equals the expected loss incurred by predicting $H(x,D)$:
  \[
    T(x,D) = \EE_{y\sim p(y|x,D)}\big [ \ell(y,H(x,D)) \big ]\:.
  \]

\noindent 2. If it rejects, $C(x,D)=0$, the penalty equals $\veps$.

The overall performance of the Bayesian reject-option predictor $Q=(H,B)$ is then evaluated by:
\begin{multline}
\label{equ:BaysianRejOptObj}
   R_\veps(H,C)=\\ \EE_{(x,D)\sim p(x,D)} \Big [C(x,D)\, T(x,D)+ \neg C(x,D)\,\veps 
       \Big ]    \:.
\end{multline}
\begin{theorem}\label{equ:BayesianRejOpt} Let $Q_B=(H_*,C_T)$ be a reject-option predictor represented by the optimal Bayesian predictor $H_*$, defined in~\equ{equ:BLOptBayes}, and the selector $C_T(x,D) = \one{ T_*(x,D) \leq \veps }$ where 
\begin{equation}
\label{equ:TotalUncertainty}
   T_*(x,D) = \EE_{y\sim p(y\mid x,D)} \big [\ell(y,H_*(x,D))\big ] \:.
\end{equation}
Then \(Q_B\) is a minimizer of \(R_{\varepsilon}(H,C)\) defined in~\eqref{equ:BaysianRejOptObj}. We refer to \(Q_B\) as the \emph{optimal Bayesian reject-option predictor}.

\end{theorem}
As we discuss below in more details, the value of $T_*(x,D)$ accounts for both aleatoric and epistemic uncertainty connected to the optimal prediction $H_*(x,D)$ and we therefore referred to it as the total uncertainty.

\begin{example}
\label{example:BayesianRejectOption}
Consider the setup from Example~\ref{example:BLregression}. The optimal Bayesian reject-option predictor $Q_B=(H_*,C_T)$ consists of $H_*(x, D)$ defined in~\equ{equ:BLpredLinReg}, and the total uncertainty~\equ{equ:TotalUncertainty}, which in this setting takes the form
\begin{equation}
\label{equ:TotalCondRiskLinReg}
  T_*(x, D) = \mathrm{Var}_{y \sim p(y \mid x, D)}[y] = \#\phi(x)^T\, \#\Sigma_m\, \#\phi(x) + v(x) \:.
\end{equation}
%
%This expression corresponds to the total variance of the predicted target.
Figure~\ref{fig:BayesianRejectOpt} illustrates the resulting optimal Bayesian reject-option predictor.
\end{example}

\section{Proposed epistemic reject-option predictor}
\label{sec:ProposedEpistemRejOption}
The goal of our framework is to construct a reject-option predictor whose accept–reject decisions are based solely on epistemic uncertainty. We formulate this predictor as the solution to a well-posed decision-making problem by departing from standard Bayesian learning, which minimizes expected loss~\eqref{equ:BaysianRejOptObj}, and instead minimizing expected regret. Regret is defined as the performance gap between the predictor inferred from data and the Bayes-optimal predictor with full knowledge of the underlying parameters. The resulting predictor abstains whenever the expected regret for a given input exceeds a specified rejection cost, which determines the maximum performance degradation, relative to the optimal predictor, that we are willing to accept.

%The goal of our framework is to construct a reject-option predictor whose accept–reject decisions are based solely on epistemic uncertainty. We formulate this predictor as the solution to a well-posed decision-making problem. We depart from standard Bayesian learning, which minimizes expected loss~\eqref{equ:BaysianRejOptObj}, and instead we minimize expected regret, defined as the performance gap between the predictor inferred from data and the Bayes-optimal predictor with full knowledge of the underlying parameters. The resulting predictor abstains whenever the expected regret for a given input exceeds a specified rejection cost. This rejection cost therefore determines the maximum performance degradation, relative to the optimal predictor, that we are willing to accept.

%The goal of our framework is to construct a reject-option predictor that bases its accept–reject decisions solely on epistemic uncertainty. We define this predictor as the solution to a well-posed decision-making problem. The central idea is to depart from standard BL, which minimizes expected loss~\equ{equ:BaysianRejOptObj}, and instead minimize expected regret, defined as the performance gap between the predictor inferred from data and the Bayes-optimal predictor with full knowledge of the underlying parameters. The predictor abstains when the expected regret for a given input data exceeds a specified rejection cost. The reject cost thus specifies how much performance decrease, relative to the optimal predictor, we are willing to still accept.

Following the BL paradigm, our aim is to construct a reject-option predictor $Q \colon \mathcal{X} \times (\mathcal{X} \times \mathcal{Y})^m \rightarrow \mathcal{Y} \cup \{ \text{reject} \}$,  which, given training data $D \in \mathcal{D}$ and an input $x \in \mathcal{X}$, either predicts $Q(x, D) \in \mathcal{Y}$ or opts to reject, $Q(x, D) = \text{reject}$. Again, we use~\equ{equ:BLRejectOptionPred} to represent $Q$ by a pair $(H,C)$, where  $H\colon \mathcal{X} \times (\mathcal{X} \times \mathcal{Y})^m \rightarrow \mathcal{Y}$ is a predictor and $C\colon \mathcal{X} \times (\mathcal{X} \times \mathcal{Y})^m \rightarrow \{0,1\}$ is a selector.

Let $\ell\colon\SY\times\SY\rightarrow\Re_+$ be a prediction loss and $\delta\in\Re_+$ a rejection cost. Then, the decision of the epistemic reject-option predictor $Q=(H,C)$ on a data sample $(x,D)\sim p(x,D)$ is penalized as follows:

\noindent 1. If it predicts, $C(x,D)=1$, the penalty equals to the expected conditional regret
  \[
    E(x,D) = \EE_{(\theta,y)\sim p(\theta,y|x,D)}\big [ \ell(y,H(x,D)) - \ell(y,h(x,\theta)) \big ]\:,
  \]
  i.e. the expected performance gap between the predictor $H(x,D)$ and the Bayes-optimal predictor $h(x,\theta)$ knowing the true parameter $\theta$.

\noindent 2. If it rejects, $C(x,D)=0$, the penalty equals $\delta$.

The overall performance of the epistemic reject-option predictor $Q=(H,C)$ is then evaluated by:
\begin{multline}
\label{equ:EpistemicRejOptObj}
   R_\delta(H,C)=\\ \EE_{(x,D)\sim p(x,D)} \Big [C(x,D)\, E(x,D)+ \neg C(x,D)\,\delta 
       \Big ]    \:.
\end{multline}

The following theorem characterizes the minimizer of $R_\delta(H,C)$, which we refer to as the optimal epistemic reject-option predictor.
\begin{theorem}\label{thm:EpistemicRejOpt} Let $Q_E=(H_*,C_E)$ be a reject-option predictor represented by the optimal Bayesian predictor $H_*$, defined in~\equ{equ:BLOptBayes}, and the selector $C_E(x,D) = \one{ E_*(x,D) \leq \delta }$
where 
\begin{multline}
  \label{equ:ConditionalRegret}
  E_*(x,D) = \\\EE_{(\theta,y)\sim p(\theta,y|x,D)}\big [ \ell(y,H_*(x,D)) - \ell(y,h(x,\theta)) \big ]\:.
\end{multline}
Then, $Q_E$ is a minimizer of $R_\delta(H,C)$ defined in~\equ{equ:EpistemicRejOptObj}. 
We refer to \(Q_E\) as the \emph{optimal epistemic reject-option predictor}.

\end{theorem}
We defer the proof to Appendix~\ref{appendix:ProofOfEpistemicTheorem}.

The expected conditional regret $E_*(x, D)$ forms a basis of the accept–reject decisions of the epistemic reject-option predictor $Q_E(x, D)$. The value of $E_*(x, D)$ admits a clear interpretation: for a given input $x$ and training set $D$, it quantifies the expected performance gap between the learned predictor and the Bayes-optimal predictor with full knowledge of the data-generating process. This makes $E_*(x, D)$ a natural measure of epistemic uncertainty, and we refer to it as such in the following. Consequently, the epistemic reject-option predictor accepts inputs for which the predictions of the learned predictor are sufficiently close, where closeness is controlled by $\delta$, to those of the theoretically optimal Bayes predictor.
\section{Summary}
\label{sec:summary}

%The aleatoric reject-option predictor $q^*(x)$ relies on the base (non-rejecting) Bayes-optiom predictor $h^*(x)$ to make predictions when a rejection is not issued. The accept-or-reject decision is based on the conditional risk $r^*(x)$, which can be interpreted as the aleatoric uncertainty when the full data-generating distribution $p(x,y)$ is known.
%
%As a theoretical construct, the aleatoric reject-option predictor assumes access to the true distribution $p(x,y)$ and neglects epistemic uncertainty. This assumption is reasonable when the amount of training data is sufficiently large. In contrast, the 

Bayesian $Q_B(x,D)$ and the epistemic $Q_E(x,D)$ reject-option predictors are designed for scenarios with finite training data $D$. Both have a similar structure: they use the optimal Bayesian predictor $H_*(x,D)$ as the base predictor, but they differ in the uncertainty measure used to decide whether to accept or reject a prediction. The Bayesian reject-option predictor relies on the total uncertainty $T_*(x,D)$, while the epistemic reject-option predictor uses the epistemic uncertainty $E_*(x,D)$. Table~\ref{tab:Summary} presents specific instances of the Bayesian and epistemic reject-option predictors for three commonly used loss functions: quadratic loss, 0/1 loss, and cross-entropy loss. 

To clarify the relationship between the uncertainties used by the two approaches, let us introduce a third quantity:
\begin{multline}
\label{equ:AleatorUncertainty}
A_*(x, D)  =  T_*(x, D) - E_*(x, D) \\
%& = & \EE_{(\theta, y) \sim p(\theta, y \mid x, D)} \left[\ell(y, h(x, \theta))\right].
 =  \EE_{\theta\sim p(\theta\mid D)}\EE_{y\sim p(y\mid x,\theta)} \left[\ell(y, h(x, \theta))\right].
\end{multline}
Recall that the conditional risk $a_*(x) = \EE_{y \sim p(y \mid x)}[\ell(y, h^*(x))]$, used by the aleatoric reject-option predictor, represents the true aleatoric uncertainty under the true data-generating distribution $p(x,y)$. By comparison, the expression for $A_*(x,D)$ in~\equ{equ:AleatorUncertainty} gives the expected conditional risk under the parameter posterior $p(\theta\mid D)$. Thus, $A_*(x,D)$ can be interpreted as the expected value of the aleatoric uncertainty in $x$ given the training set $D$, instead of the true parameters $\theta$. 
%Under standard regularity conditions, the Bayesian estimate converges to the maximum-likelihood estimate as the number of examples grows.
The decomposition (following from~\equ{equ:AleatorUncertainty})
\[ 
  T_*(x,D)=A_*(x,D)+E_*(x,D)
\]
highlights a key distinction between the proposed epistemic and Bayesian reject-option predictors. The epistemic predictor makes its accept-reject decisions based solely on epistemic uncertainty $E_*(x,D)$, whereas the Bayesian predictor uses the total uncertainty $T_*(x,D)$, which combines both aleatoric $A_*(x,D)$ and epistemic $E_*(x,D)$ components. 

\begin{example} 
Consider the setting of Example~\ref{example:BLregression}. The three types of the uncertainties decompose as follows:
\[
  \begin{array}{lrcl}
    \text{Aleatoric} & A_*(x,D) &=& v(x) \:,\\
    \text{Epistemic}  & E_*(x,D) &=& \#\phi(x)^T \#\Sigma_m\#\phi(x) \:,\\
    \text{Total}     & T_*(x,D) &=& \#\phi(x)^T\, \#\Sigma_m\, \#\phi(x) + v(x) \:.
  \end{array}
\]
Note that the aleatoric uncertainty $A_*(x,D)$ coincides with the conditional risk (i.e. true aleatoric uncertainty) $a_*(x)$, since noise variance $v(x)$ is assumed to be known in our example.  To compare the accept-reject regions of the three reject-option predictors, refer to the following figures:
\begin{itemize}[leftmargin=1.1em,itemsep=1pt]
    \item \textbf{Aleatoric:} As shown in Figure~\ref{fig:aleatoricRejectOption}, this theoretical predictor with full knowledge of $p(x,y)$ uses true aleatoric uncertainty, $a_*(x)=A(x,D)$, to define its rejection rule. Consequently, the rejection area consists of inputs with high inherent noise (i.e. high variance~$v(x)$).
    \item \textbf{Bayesian:} This predictor, illustrated in Figure~\ref{fig:BayesianRejectOpt}, bases its rejection on total uncertainty, $T_*(x,D)$. It rejects inputs that either have high aleatoric noise or are far from the training set or both happens.
    \item \textbf{Epistemic:} Shown in Figure~\ref{fig:EpistemicRejectOpt}, this predictor utilizes epistemic uncertainty, $E_*(x,D)$. Its rejection area contains inputs that are not reliably supported by the training data. Notably, this allows the predictor to accept inputs with high aleatoric uncertainty, provided its performance is comparable to that of the Bayes-optimal predictor. %This acceptable performance gap is controlled by the rejection cost~$\delta$.
    %    Notably, this allows the predictor to accept inputs with high aleatoric uncertainty, as long as its performance on them is similar to that of the Bayes-optimal predictor. The allowed performance gap is controlled via the rejection cost $\delta$.
\end{itemize}

%\begin{itemize}
%    \item The aleatoric reject-option predictor, which uses uncertainty $r^*(x)=A(x,D)$, is shown in Figure~\ref{fig:aleatoricRejectOption}. The rejection area contains inputs with high aleatoric noise.
%    \item The Bayesian reject-option predictor, based on total uncertainty $T(x,D)$, is shown in Figure~\ref{fig:BayesianRejectOpt}. The rejection area corresponds to inputs where the aleatoric noise is high and which are not covered by trainig samples.
%    \item The epistemic reject-option predictor, which uses $E(x,D)$, is shown in Figure~\ref{fig:EpistemicRejectOpt}. The rejection area contains inputs which are not relibly supported by training data. Note that the accepted inputs can have high aleatoric uncertainty, yet the learned predictor performs on these inputssimilarly to the Bayes-optimal predictor.
%\end{itemize}
\end{example}

\begin{table*}[]
    \centering
    \begin{tabular}{ll|c|c|c}
          \multicolumn{2}{c|}{Loss}& Bayesian predictor & Total uncertainty & Epistemic uncertainty\\
          \multicolumn{2}{c|}{$\ell(y,\hat{y})$} & $H_*(x,D)$        & $T_*(x,D)$          & $E_*(x,D)$ \\
       \hline%\hline
       \hline
       Squared & $(y-\hat{y})^2$ &
       $\EE_{y\sim p(y|x,D)}[y] $  &
       ${\rm Var}_{y\sim p(y|x,D)} [y]$&
       ${\rm Var}_{\theta\sim p(\theta|D)} \Big [ \EE_{y\sim p(y|x,\theta)}[y] \Big ]$\\
       \hline
       %\multicolumn{5}{l}{0/1 loss $\ell(y,\hat{y})=\leftbb y\neq \hat{y}\rightbb$} \\
       0/1 & $\one{ y\neq \hat{y}}$ & 
       $\argmax\limits_{y\in\SY} p(y|x,D)$  &
       $1-\max_{y\in\SY} p(y|x,D)$ &
       $\EE_{\theta\sim p(\theta|D)}\Big[ \max\limits_{y\in\SY} p(y|x,\theta)\Big ] - \max\limits_{y\in\SY} p(y|x,D)$ \\
       \hline
       CE & $-\log p_y$ & $p(y|x,D)$ & $\entropy{p(y|x,D)}$ &
       $\EE_{\theta\sim p(\theta|D)} \Big [ \kl{ p(y|x,\theta)\, \|\, p(y|x,D)} \Big ]$ \\       
    \end{tabular}
    \caption{Specific instances of the Bayesian (using $H_*(x,D)$ and $T_*(x,D)$) and epistemic (using $H_*(x,D)$ and $E_*(x,D)$) reject-option predictors for three commonly used loss functions: squared loss, 0/1 loss, and cross-entropy (CE) loss. The derivation assumes that $p(x,y|\theta)=p(x)\,p(y|x,\theta)$ so that the predictive distribution takes the form $p(y|x,D)=\EE_{\theta\sim p(\theta|D)}[p(y|x,\theta)]$.
    In case of the cross-entropy loss $\ell(y,\#p)=-\log p_y$, the set of target space is finite, $\SY=\{1,2,\ldots,Y\}$, while the base predictor $H\colon\SX\times(\SX\times\SY)^m\rightarrow\SP=\{\#p\in\Re_+^Y\mid \sum_{y\in\SY}p_y=1\}$ outputs a distribution over $\SY$. We use $\entropy{p(y)} = -\mathbb{E}_{y \sim p(y)}[\log_2 p(y)]$ to denote the Shannon entropy of $p(y)$, and $\kl{ p(y)\, \|\, q(y)}=\EE_{y\sim p(y)} [ \log \frac{p(y)}{q(y)}]$ to denote KL divergence between $p(y)$ and $q(y)$.}
    \label{tab:Summary}
    \vspace{-0.65cm}
\end{table*}

\section{Experiments}
In this section, we empirically validate our theoretical framework and illustrate the distinct behaviors of the reject-option predictors using a synthetic experiment. 
A conceptually similar experiment in a more realistic setting--age prediction from facial images--is presented in the Appendix~\ref{appendix:experiments} and yields comparable results. 

\vspace{-0.2cm}

\paragraph{Data} We employ a setup similar to the running example, but to increase model complexity, we use a third-degree polynomial instead of a linear regression model. For each trial, we define a ground-truth data-generating function by sampling a parameter vector $\#\theta$ from a zero-mean Gaussian prior. This vector, $\#\theta$, represents the coefficients of the cubic polynomial. The prior covariance is a diagonal matrix with a variance of $1$ for the intercept term and $0.1$ for all other coefficients. We then generate a training set $D$ of size $m$ by drawing i.i.d. samples $x \sim \mathcal{N}(0, 1)$ and generating corresponding labels $y$ from the true function with added heteroscedastic noise. Specifically, the labels are sampled according to the following distribution: $p(y|x, \#\theta) = \mathcal{N}( y\mid f(x; \#\theta), v(x))$, where $f(x; \#\theta)$ is the true polynomial function and the variance $v(x) = 0.1 + 0.04(x + 8)^2$.

\vspace{-0.2cm}

\paragraph{Important note} This synthetic generation is strictly for validation purposes to verify our theoretical findings. In practice, the proposed epistemic reject-option predictor relies only on the learned posterior and does \textbf{not} require knowledge of the ground truth for inference. Only the evaluation of the regret requires access to the data-generating process.

\vspace{-0.2cm}

%We employ similar setup as in the running example, but insteat of linear regression model we use polynomial of degree 3 to increase model complexity. For each trial, we first define a ground-truth data-generating function by sampling a parameter vector $\#\theta^*$ from a zero-mean Gaussian prior, where $\#\theta^*$ represents the coefficients of a cubic polynomial. The prior covariance is a diagonal matrix with a variance of $1$ for the intercept term and $0.1$ for all other coefficients. We then generate a training set $D$ of size $m$ by drawing i.i.d. samples $x \sim \mathcal{N}(0, 1)$ and generating corresponding labels $y$ from the true function with added heteroscedastic noise. Specifically, $p(y|x,\#\theta) = \SN( y\mid f(x; \#\theta^*), v(x))$, where $v(x) = 0.1 + 0.04 \cdot (x + 8)^2$ is the variance.

\paragraph{Compared predictors} We compare three reject-option strategies:
i) the Bayesian reject-option predictor (Thm~\ref{equ:BayesianRejOpt}), ii)
the proposed epistemic reject-option predictor (Thm~\ref{thm:EpistemicRejOpt}), and iii) a plug-in optimal aleatoric reject-option predictor using $\hat{h}(x) = h(x, \hat{\theta})$ and $\hat{r}(x) = r(x, \hat{\theta})$, where $\hat{\theta}$ is the maximum-likelihood estimate (Sec~\ref{sec:MLLearning}).
As previously shown, the inference for all the compared predictors admits a closed-form solution.

\vspace{-0.23cm}

%} (Figure \ref{fig:BayesianRejectOpt}), the proposed \textit{epistemic reject-option predictor} (Figure \ref{fig:EpistemicRejectOpt}), and the \textit{plug-in aleatoric predictor}, which first learns a predictor $\hat{h}(x)$ via maximum-likelihood estimation of $\#\theta$, and then abstains based on the known aleatoric uncertainty $r^*(x)=v(x)$.
% We then perform Bayesian inference to compute the posterior distribution over the model parameters $p(\#\theta\mid D)$. We compare three reject-option strategies: The \textit{Bayesian reject-option predictor} (Figure \ref{fig:BayesianRejectOpt}), the proposed \textit{epistemic reject-option predictor} (Figure \ref{fig:EpistemicRejectOpt}), and the \textit{plug-in aleatoric predictor}, which first learns a predictor $\hat{h}(x)$ via maximum-likelihood estimation of $\#\theta$, and then abstains based on the known aleatoric uncertainty $r^*(x)=v(x)$.

\paragraph{Evaluation metrics} 

Following standard practice in the reject-option literature, we evaluate performance using the \emph{Regret–Coverage} curve, which plots the expected regret~\eqref{equ:EpistemicRejOptObj} on non-rejected inputs against the coverage (i.e., the proportion of accepted inputs) across all possible rejection thresholds \(\epsilon\) or \(\delta\). In addition, to highlight differences with existing reject-option predictors, we also evaluate the \emph{Risk–Coverage} curve, which plots the expected (Bayesian) risk~\eqref{equ:BaysianRejOptObj} on non-rejected inputs as a function of coverage. Overall performance is summarized by the Area under the Regret–Coverage and Risk–Coverage curves, where lower values indicate better performance. Experiments are repeated for various training set sizes \(m\), and results are averaged over 5000 independent trials.

% For each learned predictor, we generate a \textit{Regret-Coverage} curve by plotting the expected regret against the coverage for all possible rejection thresholds $\epsilon$ or $\delta$. 
%Performance of the predictor is summarized by the \textit{Area under the Regret-Coverage} (AuReC) curve, where a lower area indicates better performance. The experiment is repeated with various training data sizes $m$ and the results are averaged over 3000 trials.

\begin{figure}[htb]
    \centering
    \begin{minipage}[l]{\linewidth}
        \centering
        \includegraphics[width=\linewidth]{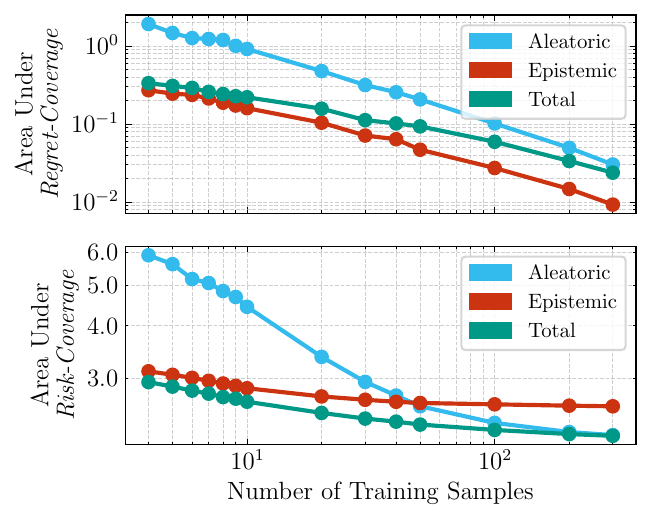}
        \captionof{figure}{
        %Performance of different reject-option predictors on a synthetic polynomial regression task. The plot shows the Area under the Regret-Coverage (AuReC) curve versus the number of training samples. Lower AuReC indicates better performance. The predictor using epistemic uncertainty consistently outperforms those using total or aleatoric uncertainty for regret minimization. Shaded regions show the central 20\% interval of results across 3000 trials.
        The performance of different reject-option predictors in a synthetic polynomial regression task. The plot shows the Area under the Regret-Coverage curve, and Area under the Risk/Coverage curve plotted against the number of training samples; Lower area is better. The results demonstrate that the proposed predictor using epistemic uncertainty consistently achieves lower regret and outperforms predictors that rely on total or aleatoric uncertainty. Results are averaged over 5000 independent trials.%Shaded areas indicate the central $20\%$ interval of outcomes across 3000 independent trials.        
        }
        \label{fig:synthetic_experiment}
    \end{minipage}
    \vspace{-0.5cm}
\end{figure}

\paragraph{Results}
The performance curves in Figure~\ref{fig:synthetic_experiment} confirm our claims. The proposed epistemic reject-option predictor consistently achieves the lowest area under the Regret–Coverage curve across all training set sizes \(m\), indicating superior performance for regret minimization. In contrast, the Bayesian reject-option predictor consistently attains the lowest area under the Risk–Coverage curve. The results further illustrate how different sources of uncertainty affect performance at different data scales. For small training sets, epistemic uncertainty is high and dominates the total uncertainty, resulting in nearly identical performance for the Bayesian and epistemic predictors. As the training set size increases, epistemic uncertainty diminishes and total uncertainty becomes increasingly dominated by the aleatoric component. Consequently, the performance of the aleatoric-only predictor obtained via ML learning converges to that of the Bayesian reject-option predictor, as expected.

\section{Existing works}
\vspace{-0.1cm}

\paragraph{Aleatoric reject-option}
The classical cost-based formulation of reject-option prediction was introduced in~\cite{Chow-RejectOpt-TIT1970}, where the optimal predictor is defined with respect to the true data distribution $p(x,y)$ and rejects based on the conditional risk, which captures aleatoric uncertainty. A common approach to learning neural-network-based reject-option predictors from data builds on Maximum-Likelihood (ML) estimation combined with the plug-in rule (cf. Sec.~\ref{sec:MLLearning}). Reject-option predictors can also be learned within the Empirical Risk Minimization (ERM) framework~\cite{Yuan-ClassiRejOpt-JMLR2010}. Both ML- and ERM-based approaches are asymptotically justified: as the training set size tends to infinity, epistemic uncertainty vanishes and the learned predictor converges to the Bayes-optimal one, rendering aleatoric uncertainty the sole driver of rejection decisions. Beyond these principled methods, a wide range of heuristic approaches has been proposed (for comprehensive overview see~\cite{Hendrickx-MLwithRejectOpt-ML2024}).

\vspace{-0.3cm}

\paragraph{Bayesian learning}
Bayesian learning (BL) (e.g.~\cite{Bishop-PRML-2006,Gelman-Bayesian-2013}) provides a principled framework for modeling predictive uncertainty by maintaining a posterior distribution over model parameters and an induced predictive distribution over the target variable, thereby capturing both aleatoric and epistemic uncertainty. Bayesian neural networks (BNNs) extend this framework to deep models by placing priors over network parameters and have been successfully applied to improve uncertainty calibration, out-of-distribution detection, and reject-option prediction~\cite{Kendal-What-NIPS2017,Lakshminarayanan-Ensembles-NIPS2017}.

\vspace{-0.15cm}

Our work builds on this line of research by establishing a decision-theoretic foundation for reject-option prediction in which epistemic uncertainty is explicitly isolated and exploited. Since exact Bayesian inference in deep networks is generally intractable, practical BNNs rely on approximations such as Monte Carlo Dropout~\cite{Gal-Dropout-ICML2016} and deep ensembles~\cite{Lakshminarayanan-Ensembles-NIPS2017}. 
%In contrast, the Bayesian linear regression models used in our running examples admit closed-form solutions, enabling a transparent demonstration of the proposed epistemic reject-option framework and a clear isolation of epistemic effects. 
%A thorough investigation of scalable implementations of the proposed framework in deep Bayesian neural networks is beyond the scope of this paper and constitutes an important direction for future work, potentially warranting a dedicated study.
%\vspace{-0.2cm}

\paragraph{Aleatoric and epistemic uncertainty representation}
\vspace{-0.1cm}
The BL produces the predictive posterior  $p(y\mid x,D)$ which allows to define the total uncertainty $T_*(x,D)$. An influential approach to decomposing total uncertainty into aleatoric and epistemic components was proposed by~\cite{Depeweg-Decompos-ICML2018}. Their method first quantifies total and aleatoric uncertainty and subsequently defines epistemic uncertainty as their difference. In the classification setting, $\SY=\{1,\ldots,Y\}$, total uncertainty is measured by the entropy of the predictive distribution $T_*(x,D)=\entropy{p(y|x,D)}$. Aleatoric uncertainty is defined as the expected entropy of the conditional distribution $p(y|x,D)$, averaged over the posterior distribution of the parameters: $A_*(x,D)=\EE_{\theta\sim p(\theta|D)}[\entropy{p(y|x,\theta)}]$. Epistemic uncertainty is then defined as $E_*(x,D) = T_*(x,D) - A_*(x,D)$, corresponding to the mutual information between $y$ and~$\theta$:
\begin{equation}
\label{equ:EntropyBaseDecomposition}
  E_*(x,D) = \EE_{\theta\sim p(\theta|x,D)} \big [{\rm D}_{\rm KL}\big (p(y|x,\theta)\, \|\, p(y|x,D)\big ) \big ] \:,
\end{equation}
A similar decomposition has been proposed for regression, $\SY=\Re$, by replacing entropy with variance. In this case, total uncertainty is defined as the variance of the predictive distribution, $T_*(x,D)={\rm Var}_{y \sim p(y \mid x, D)} [y]$ while the aleatoric uncertainty as the expected conditional variance $A_*(x,D) = \EE_{\theta \sim p(\theta \mid x,D)} \big [ {\rm Var}_{y \sim p(y \mid x, \theta)}[y] \big ]$. The resulting epistemic uncertainty, $E_*(x,D) = T_*(x,D) - A_*(x,D)$, corresponds to the variance of the conditional mean w.r.t. to posterior over model parameters:
\begin{equation}
\label{equ:VarianceBaseDecomposition}
  E_*(x,D) = {\rm Var}_{\theta\sim p(\theta|D)} \big [ \EE_{y\sim p(y|x,\theta)}[y] \big ] \:.
\end{equation}
More recently,~\cite{Hofman-QuantifAleatoricEpistemic-Arxiv2024} proposed a general framework for quantifying aleatoric and epistemic uncertainty based on proper scoring rules, showing that both the entropy-based and variance-based decompositions arise as special cases of this more general approach.

\vspace{-0.15cm}

The entropy-based~\eqref{equ:EntropyBaseDecomposition} and variance-based~\eqref{equ:VarianceBaseDecomposition} definitions of epistemic uncertainty coincide with our definition of conditional regret~\eqref{equ:ConditionalRegret} when instantiated with the cross-entropy or squared loss, respectively (see Table~\ref{tab:Summary}). While this leads to identical expressions, the underlying methodology is fundamentally different. Rather than postulating an uncertainty decomposition \emph{a priori}, we derive epistemic uncertainty by first defining the optimal epistemic reject-option predictor as the minimizer of the expected regret~\eqref{equ:EpistemicRejOptObj}, and then showing that the resulting optimal accept–reject decision is governed by the conditional regret $E_*(x,D)$.

\vspace{-0.15cm}

Our formulation therefore provides a decision-theoretic justification for widely used measures of epistemic uncertainty that were previously introduced from different principles and applied across diverse tasks. Notably, the definitions in~\eqref{equ:EntropyBaseDecomposition} and~\eqref{equ:VarianceBaseDecomposition} have proven effective in applications such as active learning, reinforcement learning, and credible interval estimation~\cite{Depeweg-Decompos-ICML2018,Wang-EpistQuant-CVPR2024}. However, to the best of our knowledge, these measures have not been employed for constructing reject-option predictors.

\vspace{-0.15cm}

We also note that some recent work has questioned the validity of the entropy-based uncertainty decomposition proposed in~\cite{Depeweg-Decompos-ICML2018}. For example,~\cite{Wimmer-Proper-CUAI2023} introduce an axiomatic framework for epistemic and aleatoric uncertainty and show that entropy-based decompositions violate certain axioms. Our contribution offers a complementary perspective: we demonstrate that the epistemic uncertainty $E_*(x, D)$ defined in~\eqref{equ:EntropyBaseDecomposition} is the optimal uncertainty measure for epistemic reject-option prediction under the cross-entropy loss. This suggests that the axiomatic notion of uncertainty in~\cite{Wimmer-Proper-CUAI2023} may not be fully aligned with the objectives and decision-theoretic requirements of the reject-option setting.

\vspace{-0.1cm}

\section{Conclusions}
This paper introduces the concept of epistemic reject-option predictor, which abstains from making predictions when epistemic uncertainty, i.e. uncertainty due to limited data, is high. Unlike standard reject-option approach focusing solely on aleatoric uncertainty, or Bayesian reject-option predictors based on total uncertainty, our framework is, to our knowledge, the first to systematically identify inputs for which training data is insufficient to support well-informed predictions. 

\vspace{-0.1cm}

We formalize this approach by minimizing expected regret, defined as the performance gap between the learned predictor and the Bayes-optimal predictor with full knowledge of the data-generating process. The resulting optimal epistemic reject-option predictor abstains whenever the conditional regret for a given input exceeds a user-specified rejection cost, which controls the maximum performance degradation we are willing to accept.

\vspace{-0.05cm}

Notably, widely used measures of epistemic uncertainty in Bayesian neural networks, such as entropy- and variance-based scores, correspond to the conditional regret under specific loss functions. While epistemic uncertainty has been leveraged in various areas, we present, to our knowledge, its first principled application to reject-option prediction.

%To provide an empirical validation, we apply our framework to regression models for which all relevant quantities can be computed analytically, using simulated data that enable a direct evaluation of expected regret. In addition, we consider a more realistic setting--age prediction from facial images--and observe consistent behavior (c.f. Appendix~\ref{appendix:experiments}). The experimental results support our theoretical findings. Investigating scalable implementations of the proposed framework in more complex models, such as deep Bayesian neural networks, represents an important direction for future work. However, we see as the main contribution formulating the theoretical framework that is the necessary first step for developing practical prediction models able to identify inputs on which their predictions are supported by data.

\vspace{-0.05cm}
To provide an empirical validation, we apply our framework to regression models for which all relevant quantities can be computed analytically, using simulated data that enable a direct evaluation of expected regret. In addition, we consider a more realistic setting—age prediction from facial images—and observe consistent behavior (cf. Appendix~\ref{appendix:experiments}). Investigating scalable implementations of the proposed framework in complex models, such as deep Bayesian neural networks, represents an important direction for future work. We emphasize that the primary contribution of this work lies in formulating the underlying theoretical framework, which constitutes a necessary first step toward developing practical prediction models capable of identifying inputs for which their predictions are supported by the available data.

%by extending standard Bayesian learning: rather than minimizing expected loss, we minimize expected regret, defined as the performance gap between the learned predictor and the Bayes-optimal predictor with full knowledge of the data-generating process. The resulting optimal epistemic predictor makes accept–reject decisions based on the conditional regret, which naturally quantifies epistemic uncertainty. 
%The epistemic reject-option predictor abstains when the regret for a given input exceeds a specified rejection cost.

%We demonstrate the framework on simulated data using a simple regression model where all key quantities are analytically tractable. Evaluating the proposed framework on real-world tasks requires intensive computational approximations, which we leave for future work.

\section*{Impact Statements}
This paper presents work whose goal is to advance the field of machine learning. There are many potential societal consequences of our work, none of which we feel must be specifically highlighted here.

\bibliography{main.bib}

@book{Wackerly-2002,
  title={Mathematical Statistics with Applications},
  author={Wackerly, Dennis D. and Mendenhall, William and Scheaffer, Richard L.},
  year={2002},
  edition={6th},
  publisher={Duxbury Press},
  address={Boston, MA}
}

@ARTICLE{Hullermeier-Intro-ML2021,
  author={Hullermeier, E. and Waegeman, W.},
  journal={Machine Learning}, 
  title={Aleatoric and epistemic uncertainty in machine learning: An introduction to concepts and methods}, 
  year={2021},
  volume={3},
  number={110},
  pages={457-506},
}

@book{ Gelman-Bayesian-2013,
  address = {Boca Raton, Florida},
  author = {Gelman, Andrew and Carlin, John B. and Stern, Hal S. and Dunson, David B. and Vehtari, Akti and Rubin, Donald B.},
  edition = {Third},
  title = {Bayesian Data Analysis},
  year = 2013
}

@book{Bishop-PRML-2006,
author = {Bishop, Christopher M.},
title = {Pattern Recognition and Machine Learning (Information Science and Statistics)},
year = {2006},
publisher = {Springer-Verlag},
address = {Berlin, Heidelberg}
}

@misc{Hofman-QuantifAleatoricEpistemic-Arxiv2024,
      title={Quantifying Aleatoric and Epistemic Uncertainty with Proper Scoring Rules}, 
      author={Paul Hofman and Yusuf Sale and Eyke Hüllermeier},
      year={2024},
      eprint={2404.12215},
      archivePrefix={arXiv},
      primaryClass={cs.LG},
}

@InProceedings{Wimmer-Proper-CUAI2023,
  title = 	 {Quantifying aleatoric and epistemic uncertainty in machine learning: Are conditional entropy and mutual information appropriate measures?},
  author =       {Wimmer, Lisa and Sale, Yusuf and Hofman, Paul and Bischl, Bernd and H\"ullermeier, Eyke},
  booktitle = 	 {Proceedings of the Conference on Uncertainty in Artificial Intelligence},
  pages = 	 {2282--2292},
  year = 	 {2023},
  volume = 	 {216},
}

@inproceedings{Kendal-What-NIPS2017,
 author = {Kendall, Alex and Gal, Yarin},
 booktitle = {Advances in Neural Information Processing Systems},
 title = {What Uncertainties Do We Need in Bayesian Deep Learning for Computer Vision?},
 volume = {30},
 year = {2017}
}

@INPROCEEDINGS{Wang-EpistQuant-CVPR2024,
 author = { Wang, Hanjing and Ji, Qiang },
 booktitle = {Conference on Computer Vision and Pattern Recognition},
 title = {{ Epistemic Uncertainty Quantification for Pretrained Neural Networks }},
 year = {2024},
 volume = {},
 pages = {11052-11061},
 }

@InProceedings{Depeweg-Decompos-ICML2018,
  title = 	 {Decomposition of Uncertainty in {B}ayesian Deep Learning for Efficient and Risk-sensitive Learning},
  author =       {Depeweg, Stefan and Hernandez-Lobato, Jose-Miguel and Doshi-Velez, Finale and Udluft, Steffen},
  booktitle = 	 {Proceedings of the International Conference on Machine Learning},
  pages = 	 {1184--1193},
  year = 	 {2018},
  volume = 	 {80},
}

@inproceedings{Lakshminarayanan-Ensembles-NIPS2017,
  author = {Lakshminarayanan, Balaji and Pritzel, Alexander and Blundell, Charles},
  booktitle = {Advances in Neural Information Processing Systems},
  title = {Simple and Scalable Predictive Uncertainty Estimation using Deep Ensembles},
  volume = {30},
  year = {2017}
}

@InProceedings{Gal-Dropout-ICML2016,
  title = 	 {Dropout as a Bayesian Approximation: Representing Model Uncertainty in Deep Learning},
  author = 	 {Gal, Yarin and Ghahramani, Zoubin},
  booktitle = 	 {Proceedings of International Conference on Machine Learning},
  pages = 	 {1050--1059},
  year = 	 {2016},
  volume = 	 {48},
  series = 	 {Proceedings of Machine Learning Research},
  month = 	 {20--22 Jun},
}

@article{Yuan-ClassiRejOpt-JMLR2010,
   title={Classification Methods with Reject Option Based on Convex Risk Minimization},
   author={M. Yuan and M. Wegkamp},
   journal={Journal of Machine Learning Research},
   year={2010},
   volume={11},
   pages={111-130},
}

@article{Hendrickx-MLwithRejectOpt-ML2024,
    author = {K. Hendrickx and L. Perini and D. Van der Plas and W. Meert and J. Davis} ,
    title = {Machine learning with reject option: a survey},
    journal = {Machine Learning},
    year = {2024},
    volume = {113},
    pages = {3073--3110},
}

@article{ Chow-RejectOpt-TIT1970,
    author  = {C. Chow}, 
    title   = {On optimum recognition error and reject tradeoff},
    journal = {IEEE Transactions on Information Theory},
    volume  = {16},
    number  = {1}, 
    pages   = {41–46},
    year    = {1970},
}

@inproceedings{Karras2021,
  author = {Tero Karras and Miika Aittala and Samuli Laine and Erik H\"ark\"onen and Janne Hellsten and Jaakko Lehtinen and Timo Aila},
  title = {Alias-Free Generative Adversarial Networks},
  booktitle = {Proc. NeurIPS},
  year = {2021}
}

@InProceedings{Zheng_2022_CVPR,
    author    = {Zheng, Yinglin and Yang, Hao and Zhang, Ting and Bao, Jianmin and Chen, Dongdong and Huang, Yangyu and Yuan, Lu and Chen, Dong and Zeng, Ming and Wen, Fang},
    title     = {General Facial Representation Learning in a Visual-Linguistic Manner},
    booktitle = {Proceedings of the IEEE/CVF Conference on Computer Vision and Pattern Recognition (CVPR)},
    month     = {June},
    year      = {2022},
    pages     = {18697-18709}
}
\bibliographystyle{icml2026}

%%%%%%%%%%%%%%%%%%%%%%%%%%%%%%%%%%%%%%%%%%%%%%%%%%%%%%%%%%%%%%%%%%%%%%%%%%%%%%%
%%%%%%%%%%%%%%%%%%%%%%%%%%%%%%%%%%%%%%%%%%%%%%%%%%%%%%%%%%%%%%%%%%%%%%%%%%%%%%%
% APPENDIX
%%%%%%%%%%%%%%%%%%%%%%%%%%%%%%%%%%%%%%%%%%%%%%%%%%%%%%%%%%%%%%%%%%%%%%%%%%%%%%%
%%%%%%%%%%%%%%%%%%%%%%%%%%%%%%%%%%%%%%%%%%%%%%%%%%%%%%%%%%%%%%%%%%%%%%%%%%%%%%%
\newpage
\appendix
\onecolumn
\section{Experiments}
\label{appendix:experiments}
In this section, we empirically validate our theoretical framework. A key challenge in evaluating reject-option strategies based on regret is that the calculation of \textit{True Regret} requires access to the true data-generating process $p(y|x, \theta_*)$, which is unknown for real-world datasets.

To address this, we construct a controlled high-dimensional setup that allows us to generate realistic image data while maintaining full access to the ground-truth data-generating process. This enables the exact computation of the Bayes-optimal prediction, the true regret and the true aleatoric uncertainty for every sample, allowing us to benchmark the proposed epistemic reject-option against standard baselines.

\paragraph{Important Note:} This synthetic generation is strictly for validation purposes to verify \Cref{thm:EpistemicRejOpt}. In practice, the proposed Epistemic Reject-Option predictor ($Q_E$) relies only on the learned posterior and does \textbf{not} require knowledge of the ground truth for inference. Only the evaluation of the regret requires access to the data-generating process $p(y|x, \theta_*)$.

\subsection{Experimental Setup}
\paragraph{Data Generation} We define the input distribution $p(x)$ using StyleGAN3 \cite{Karras2021} trained on the Flickr-Faces-HQ Dataset (FFHQ) dataset, which generates high-quality synthetic face images. To define the conditional distribution $p(y|x)$, we create a \textit{Ground-Truth Generator} as follows:
\begin{enumerate}
    \item We train a ViT-B-16 age estimation model on all publicly available age estimation datasets using cross-entropy loss.
    \item We distill the model into a ResNet-50 regression network. This network acts as the oracle: for any input $x$, it outputs a ground-truth mean $\mu_*(x)$ and variance $\sigma^{2}_*(x)$.
    \item Our data generation process is thus defined as $y \mid x \sim \mathcal{N}(\mu_*(x), \sigma_*^{2}(x))$.
\end{enumerate}

For our experiments, we sample a test set of $100,000$ images from $p(x)$. Training sets of varying sizes $m$ are generated by sampling $x \sim p(x)$ and labeling them by sampling a single target $y$ from the $p(y|x)=\mathcal{N}(\mu_*(x), \sigma^{2}_*(x))$.

\paragraph{Models and Training} We employ Bayesian Linear Regression (BLR) as the learning algorithm, implemented as a Gaussian Process with a linear kernel. The learner operates on a fixed feature representation $\phi(x)$ of the input image. We compare three reject-option strategies:
\begin{enumerate}
    \item \textbf{Aleatoric:} Rejects inputs with high irreducible noise using either the true $A_*(x, D)=\sigma_*^2(x)$ or its estimate.
    \item \textbf{Total:} Rejects based on $T_*(x, D)$, the variance of the predictive distribution.
    \item \textbf{Epistemic:} Rejects based on the expected conditional regret $E_*(x, D)$.
\end{enumerate}
    
We employ the Area under the Regret-Coverage Curve and the Area under the Risk-Coverage Curve as our metrics. Risk and regret are calculated using the squared loss.

\subsection{Validation on Well-Specified Models}
We first investigate the setting where the learner is well-specified, meaning the ground-truth function lies within the hypothesis class of the learner. To achieve this, we use the penultimate layer features of the Ground-Truth Generator, the ResNet-50 used to define $p(y|x)$, as the input features $\#\phi(x)\in\mathbb{R}^d$, $d=2048$, for the Bayesian Linear Regressor.

\begin{figure}[htb]
    \centering
    \begin{minipage}[t]{0.48\linewidth}
        \centering
        \includegraphics[width=\linewidth]{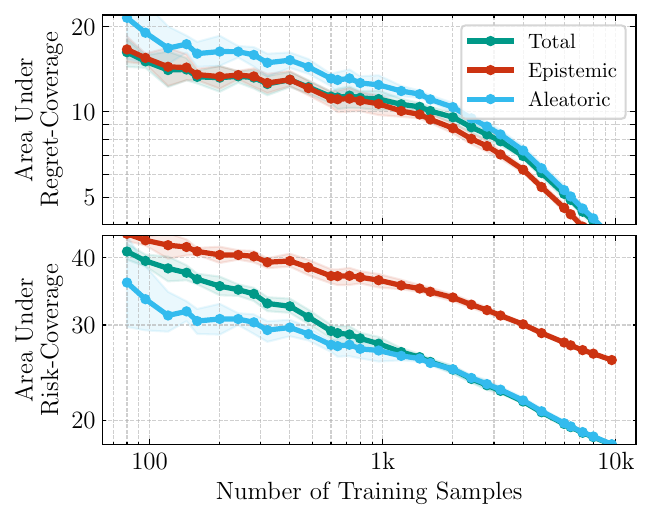}
        \caption{Performance comparison in the well-specified setting using oracle aleatoric variance. The top row shows the Area Under Regret-Coverage, and the bottom row shows the Area Under Risk-Coverage. As predicted by the theory, the Epistemic reject-option (red) consistently minimizes regret, while the Aleatoric reject-option (blue) minimizes risk. The Total uncertainty (green) acts as an interpolation: in the low-data regime, it is dominated by epistemic uncertainty and follows the red curve; as sample size increases, epistemic uncertainty vanishes, and it converges to the aleatoric behavior (blue curve).}
        \label{fig:oracle_resnet_gp_linear}
    \end{minipage}
    \hfill
    \begin{minipage}[t]{0.48\linewidth}
        \centering
        \includegraphics[width=\linewidth]{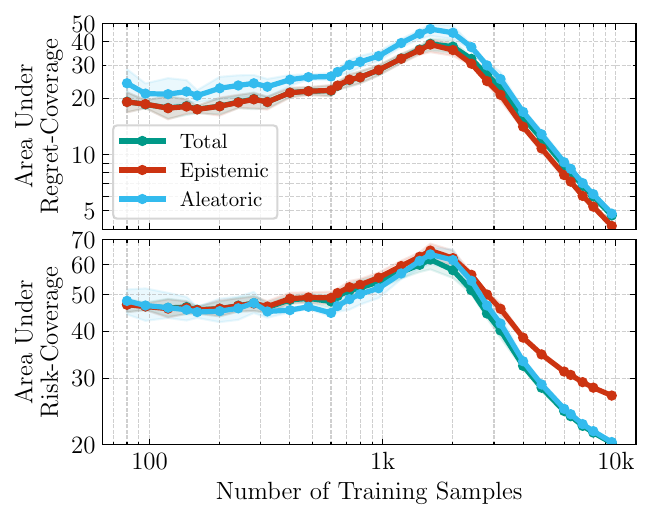}
        \caption{Performance comparison in the well-specified setting using learned aleatoric variance. Here, the noise parameter $\sigma^2$ is estimated via MLE. A characteristic peak in error appears around the interpolation threshold ($|D| \approx d = 2048$). In the under-determined regime ($|D| < 2048$), where the noise estimate often collapses to zero, the Epistemic reject-option (red) remains robust, consistently achieving the lowest regret. This confirms that the posterior variance $E_*(x, D)$ can serve as a proxy for model error even when noise estimation is unstable.}
        \label{fig:resnet_gp_linear}
    \end{minipage}
\end{figure}

\paragraph{Oracle Aleatoric Variance} Initially, we provide the learner with the true aleatoric variance $\sigma^{2}_*(x)$ to isolate the behavior of the uncertainty estimates. \Cref{fig:oracle_resnet_gp_linear} displays the performance. The proposed Epistemic strategy (red) consistently achieves the lowest Area under Regret Coverage across all sample sizes, empirically confirming that $E_*(x, D)$ is the optimal ranking function for minimizing regret. Conversely the Aleatoric strategy (blue) minimizes Risk. This confirms the theoretical distinction: the Epistemic predictor accepts noisy samples if the model has learned the mean accurately (low regret, high risk), whereas the Aleatoric predictor rejects them.

\paragraph{Learned Aleatoric Variance} In a realistic setting, $\sigma^{2}_*(x)$ is unknown. We train separate linear layers to estimate the mean $\hat{\mu}(x)$ and noise variance $\hat{\sigma}^2(x)$ via Maximum Likelihood Estimation and use this variance estimate to calibrate the BLR. \Cref{fig:resnet_gp_linear} presents the results. We observe a distinct behavior around $|D| \approx 2048$ samples ($|D| \approx d$). In the under-determined regime ($|D| < d$), the linear model is capable of perfectly interpolating the training data. Consequently, the MLE estimate for the aleatoric variance $\hat{\sigma}^2$ collapses toward zero, effectively leading the learner to treat the data as noiseless. As $|D|$ approaches $d$, the model transitions from interpolation to regression, resulting in the characteristic \textit{double descent} peak in risk and regret due to the ill-conditioning of the covariance matrix. Notably, despite the variance estimate collapsing in the low-data regime and the spectral instability at the interpolation threshold, the Epistemic reject-option maintains the property of minimizing regret. This demonstrates that the posterior variance of the weights $E_*$(x, D) remains a robust signal for model error, even when the aleatoric noise estimate is unreliable.

\subsection{Limitations under Model Misspecification}
We further investigate a \textit{misspecified} setting where the ground-truth function lies outside the hypothesis class. We replicate the experimental setup using FaRL features \cite{Zheng_2022_CVPR}, rendering the true function $\mu_*(x)$ non-linear with respect to the linear Bayesian predictor. In this regime, as $|D| \to \infty$, the parameter posterior concentrates around a single linear approximation. Consequently, the epistemic uncertainty vanishes, yet the regret converges to a non-zero constant corresponding to the approximation error (bias). This reveals a fundamental property: the proposed epistemic reject-option specifically targets \textit{estimation error} (finite data) rather than \textit{approximation error} (limited capacity). Notably, preliminary investigations suggest this behavior is not unique to the linear setting; we observed similar results with Gaussian Processes with RBF kernel and Deep approximations (MC Dropout, Deep Ensembles), highlighting an open challenge for the field.

%%%%%%%%%%%%%%%%%%%%%%%%%%%%%%%%%%%%%%%%%%%%%%%%%%%%%%%%%%%%%%%%%
\section{Proof of Theorem~\ref{equ:BayesianRejOpt}}
\label{appendix:ProofBayesianRejOpt}
%%%%%%%%%%%%%%%%%%%%%%%%%%%%%%%%%%%%%%%%%%%%%%%%%%%%%%%%%%%%%%%%%

The proof of Theorem~\ref{equ:BayesianRejOpt} is standard; we include it here for completeness and to keep the paper self-contained.

\begin{proof}The goal is to determine a base predictor
\(H \colon \SX \times (\SX \times \SY)^m \rightarrow \SY\)
and a selector \(C \colon \SX \times (\SX \times \SY)^m \rightarrow \{0,1\}\),
which together define a reject-option predictor
\(Q = (H,C)\) (cf.~\eqref{equ:BLRejectOptionPred}),
that minimizes the objective
\[
R_{\varepsilon}(H,C)
=
\EE_{(x,D)\sim p(x,D)}
\Big[
C(x,D)\,T(x,D)
+
\bigl(1 - C(x,D)\bigr)\,\varepsilon
\Big],
\]
where
\[
T(x,D)
=
\EE_{y\sim p(y \mid x,D)}
\bigl[
\ell\bigl(y, H(x,D)\bigr)
\bigr].
\]
Since the expectation decomposes over pairs \((x,D)\), the optimization can be carried out independently for each \((x,D)\). We therefore fix an arbitrary \((x,D)\) and proceed in two steps.

\medskip
\noindent
\textbf{Step 1: Optimization w.r.t. \(H(x,D)\).}
Assume that \(C(x,D)\) is fixed.
The contribution of \((x,D)\) to \(R_{\varepsilon}(H,C)\) is then
\[
C(x,D)\,
\EE_{y\sim p(y \mid x,D)}
\bigl[
\ell(y,H(x,D))
\bigr]
+
\bigl(1 - C(x,D)\bigr)\,\varepsilon .
\]
Since the second term does not depend on \(H(x,D)\), the optimal predictor is given by
\[
H_*(x,D)
=
\argmin_{\hat y \in \SY}
\EE_{y\sim p(y \mid x,D)}
\bigl[
\ell(y,\hat y)
\bigr],
\]
which coincides with the optimal Bayesian predictor defined in~\eqref{equ:BLOptBayes}.

\medskip
\noindent
\textbf{Step 2: Optimization w.r.t. \(C(x,D)\).}
Now fix \(H(x,D) = H_*(x,D)\).
The remaining optimization problem reduces to
\[
\argmin_{\hat c \in \{0,1\}}
\Big[
\hat c\,T_*(x,D)
+
(1-\hat c)\,\varepsilon
\Big],
\]
where
\(
T_*(x,D)
=
\EE_{y\sim p(y \mid x,D)}
[
\ell(y, H_*(x,D))
]
\).
The optimal selector is therefore
\[
C_T(x,D)
=
\begin{cases}
1, & \text{if } \;\; T_*(x,D) \le \varepsilon, \\
0, & \text{if } \;\;T_*(x,D) > \varepsilon .
\end{cases}
\]
Equivalently,
\(C_T(x,D) = \one{T_*(x,D) \le \varepsilon}\).

Combining the two steps yields the claimed form of the optimal Bayesian reject-option predictor, $Q_B=(H_*,C_T)$.
\end{proof}

%%%%%%%%%%%%%%%%%%%%%%%%%%%%%%%%%%%%%%%%%%%%%%%%%%%%%%%%%%%%%%%%%
\section{Proof of Theorem~\ref{thm:EpistemicRejOpt}}
%%%%%%%%%%%%%%%%%%%%%%%%%%%%%%%%%%%%%%%%%%%%%%%%%%%%%%%%%%%%%%%%%
\label{appendix:ProofOfEpistemicTheorem}

The proof of Theorem~\ref{thm:EpistemicRejOpt} follows the same structure as the proof of Theorem~\ref{equ:BayesianRejOpt} given in Appendix~\ref{appendix:ProofBayesianRejOpt}. The key difference lies in the objective, which is based on expected regret rather than expected risk.

\begin{proof}
The goal is to determine a base predictor
\(H \colon \SX \times (\SX \times \SY)^m \rightarrow \SY\)
and a selector
\(C \colon \SX \times (\SX \times \SY)^m \rightarrow \{0,1\}\),
which together define a reject-option predictor
\(Q = (H,C)\) (cf.~\eqref{equ:BLRejectOptionPred}),
that minimizes the objective
\[
R_{\delta}(H,C)
=
\EE_{(x,D)\sim p(x,D)}
\Big[
C(x,D)\,E(x,D)
+
\bigl(1 - C(x,D)\bigr)\,\delta
\Big],
\]
where
\[
E(x,D)
=
\EE_{(\theta,y)\sim p(\theta,y \mid x,D)}
\bigl[
\ell\bigl(y, H(x,D)\bigr)
-
\ell\bigl(y, h(x,\theta)\bigr)
\bigr].
\]

Since the expectation decomposes over pairs \((x,D)\), the optimization can be carried out independently for each \((x,D)\).
We therefore fix an arbitrary \((x,D)\) and proceed in two steps.

\medskip
\noindent
\textbf{Step 1: Optimization w.r.t. \(H(x,D)\).}
Assume that \(C(x,D)\) is fixed.
The contribution of \((x,D)\) to \(R_{\delta}(H,C)\) is then
\[
C(x,D)\,
\EE_{(\theta,y)\sim p(\theta,y \mid x,D)}
\bigl[
\ell(y,H(x,D)) - \ell(y,h(x,\theta))
\bigr]
+
\bigl(1 - C(x,D)\bigr)\,\delta .
\]
Since the second term and the \(\EE_{\theta,y}[\ell(y,h(x,\theta))]\) do not independent of \(H(x,D)\),
the optimal predictor is given by
\[
H_*(x,D)
=
\argmin_{\hat y \in \SY}
\EE_{(\theta,y)\sim p(\theta,y \mid x,D)}
\bigl[
\ell(y,\hat y)
\bigr]
=
\argmin_{\hat y \in \SY}
\EE_{y\sim p(y \mid x,D)}
\bigl[
\ell(y,\hat y)
\bigr].
\]
Thus, \(H_*(x,D)\) coincides with the optimal Bayesian predictor defined in~\eqref{equ:BLOptBayes}.

\medskip
\noindent
\textbf{Step 2: Optimization w.r.t. \(C(x,D)\).}
Now fix \(H(x,D) = H_*(x,D)\).
The remaining optimization problem reduces to
\[
\argmin_{\hat c \in \{0,1\}}
\Big[
\hat c\,E_*(x,D)
+
(1-\hat c)\,\delta
\Big],
\]
where
\(
E_*(x,D)
=
\EE_{(\theta,y)\sim p(\theta,y \mid x,D)}
\bigl[
\ell(y,H_*(x,D)) - \ell(y,h(x,\theta))
\bigr]
\).
The optimal selector is therefore
\[
C_E(x,D)
=
\begin{cases}
1, & \text{if }\;\; E_*(x,D) \le \delta, \\
0, & \text{if }\;\; E_*(x,D) > \delta .
\end{cases}
\]
Equivalently,
\(C_E(x,D) = \one{E_*(x,D) \le \delta}\).

Combining the two steps yields the claimed form of the optimal epistemic reject-option predictor.

\end{proof}

%%%%%%%%%%%%%%%%%%%%%%%%%%%%%%%%%%
\section{Derivation of Epistemic Uncertainty for Specific Loss Functions}
%%%%%%%%%%%%%%%%%%%%%%%%%%%%%%%%%%

Table~\ref{tab:Summary} summarizes special instances of the Bayesian predictor \(H_*(x,D)\), the total uncertainty \(T_*(x,D)\), and the epistemic uncertainty \(E_*(x,D)\) for three commonly used loss functions: squared loss, 0/1 loss, and cross-entropy loss. Throughout the paper, we assume that the parametric model of the data factorizes as
\[
p(x,y \mid \theta) = p(x)\,p(y \mid x,\theta),
\]
so that the predictive distribution takes the form
\[
p(y \mid x,D)
=
\EE_{\theta \sim p(\theta \mid D)}
\bigl[
p(y \mid x,\theta)
\bigr].
\]
This assumption allows several key quantities to be expressed in a simplified and unified manner.

The Bayesian predictor is given by
\[
H_*(x,D)
=
\argmin_{\hat y \in \SY}
\EE_{\theta \sim p(\theta \mid D)}
\Big[
\EE_{y \sim p(y \mid x,\theta)}
\bigl[
\ell(\hat y,y)
\bigr]
\Big].
\]

The total uncertainty associated with this predictor is
\[
T_*(x,D)
=
\EE_{y \sim p(y \mid x,D)}
\bigl[
\ell\bigl(H_*(x,D),y\bigr)
\bigr]
=
\EE_{\theta \sim p(\theta \mid D)}
\Big[
\EE_{y \sim p(y \mid x,\theta)}
\bigl[
\ell\bigl(H_*(x,D),y\bigr)
\bigr]
\Big].
\]

The epistemic uncertainty, measured via the conditional regret, is defined as
\[
E_*(x,D)
=
\EE_{\theta \sim p(\theta \mid D)}
\Big[
\EE_{y \sim p(y \mid x,\theta)}
\bigl[
\ell\bigl(H_*(x,D),y\bigr)
-
\ell\bigl(h(x,\theta),y\bigr)
\bigr]
\Big].
\]

Finally, recall that the Bayes-optimal predictor corresponding to a fixed parameter value \(\theta\) is given by
\[
h(x,\theta)
=
\argmin_{\hat y \in \SY}
\EE_{y \sim p(y \mid x,\theta)}
\bigl[
\ell(\hat y,y)
\bigr].
\]

%%%%%%%%%%%%%%%%%%%%%%%%%%%%%%%%%%
%%%%%%%%%%%%%%%%%%%%%%%%%%%%%%%%%%
\subsection{Squared loss}

Let $\ell(y,\hat{y})=(y-\hat{y})^2$ denote the  squared loss defined over the target space $\SY=\Re$. 

The Bayes-optimal predictor given the true parameter $\theta$ reads:
\[ h_*(x,\theta) = 
  \argmin_{\hat{y}\in\SY}\EE_{y\sim p(y\mid x,\theta)} (y-\hat{y})^2 = \EE_{y\sim p(y|x,\theta)}\big[ y\big ] = \mu_{y|x,\theta}\:.      
\]
The Bayesian predictor:
\begin{eqnarray*}
   H_*(x,D) &= &
\argmin_{\hat{y}\in\SY} \EE_{\theta\sim p(\theta|D)} \Big [\EE_{y\sim p(y|x,\theta)}\big[ (\hat{y}-y)^2 \big ]\Big ]    \\
   &=& \argmin_{\hat{y}\in\SY} \EE_{\theta\sim p(\theta|D)} \Big [
   \hat{y}^2-2\,\hat{y}\,\EE_{y\sim p(y|x,\theta)}[y] + 
   \EE_{y\sim p(y|x,\theta)}[y^2] \Big ]
   \\
  &= &\EE_{\theta\sim p(\theta|D)}\Big[\EE_{y\sim p(y|x,\theta)}[y] \Big ] =  \mu_{y|x,D}\:.
\end{eqnarray*}
The total uncertainty:
\begin{eqnarray*}
   T_*(x,D) &= &\EE_{\theta\sim p(\theta|D)} \Big [\EE_{y\sim p(y|x,\theta)}\big[ (\mu_{y|x,D}-y)^2 \big ]\Big ] = {\rm Var}_{y\sim p(y|x,D)}\big [ y\big ]\:,
\end{eqnarray*}
The epistemic uncertainty:
\begin{eqnarray*}
   E_*(x,D) &= &\EE_{\theta\sim p(\theta|D)} \bigg [\EE_{y\sim p(y|x,\theta)}\Big[ (H_*(x,D)-y)^2-(h(x,\theta)-y)^2 \Big ]\bigg ]\\
   &= &\EE_{\theta\sim p(\theta|D)} \bigg [\EE_{y\sim p(y|x,\theta)}\Big[ y^2-2\,y\,\mu_{y|x,D} + \mu_{y|x,D}^2-y^2+
   2\,y\,\mu_{y|x,\theta}+  \mu_{y|x,\theta}^2 \Big ]\bigg ]\\   
   &= &\EE_{\theta\sim p(\theta|D)} \bigg [\EE_{y\sim p(y|x,\theta)}\Big[ 2\,y\,(\mu_{y|x,\theta} - \mu_{y|x,D})+
   \mu_{y|x,D}^2-  \mu_{y|x,\theta}^2 \Big ]\bigg ]\\   
   &= &\EE_{\theta\sim p(\theta|D)} \bigg [ -2\,\mu_{y|x,\theta}\,\mu_{y|x,D} + \mu_{y|x,D}^2+
   \mu_{y|x,\theta}^2\bigg ]\\   
   &= &\EE_{\theta\sim p(\theta|D)} \bigg [ \Big(\mu_{y|x,D}-\mu_{y|x,\theta}\Big)^2 \bigg ]\\   
   &=& 
   {\rm Var}_{\theta\sim p(\theta|D)} \Big [ \EE_{y\sim p(y|x,\theta)}[y] \Big ] \:.   
   %{\rm Var}_{\theta\sim p(\theta|D)} \Big [ \mu_{y|x,\theta} \Big ] \:.
\end{eqnarray*}

%%%%%%%%%%%%%%%%%%%%%%%%%%%%%%%%%%
%%%%%%%%%%%%%%%%%%%%%%%%%%%%%%%%%%
\subsection{0/1-loss}

Let $\SY=\{1,2,\ldots,Y\}$ be a finite set of labels. Let $\ell(y,\hat{y})=\one{ y\neq \hat{y}}$ be the 0/1-loss equal to $1$ if $y\neq\hat{y}$ and $0$ otherwise.

The Bayes-optimal predictor given the true parameter $\theta$ outputs:
\[
  h_*(x,\theta) = 
  \argmin_{\hat{y}\in\SY} \EE_{y\sim p(y|x,\theta)}\big [\one{ y\neq \hat{y}} \big ] = \argmax_{y\in\SY} p(y|x,\theta )\:.  %\argmin_{\hat{y}\in\SY} \sum_{y\in\SY} p(y|x,\theta)\, \leftbb y\neq \hat{y}\rightbb = \argmax_{y\in\SY} p(y|x,\theta )\:.    
\]
The Bayesian predictor:
\[
   H_*(x,D) = 
  \argmin_{\hat{y}\in\SY} \EE_{y\sim p(y|x,D)}\big [\one{ y\neq \hat{y} }\big ] =  
  \argmax_{y\in\SY} p(y|x,D ) =
  \argmax_{y\in\SY} \EE_{\theta\sim p(\theta|D)}\big[p(y|x,\theta )\big ]\:.  
\]
The total uncertainty:
\[
 T_*(x,D)= \EE_{y\sim p(y|x,D)}\Big[ \one{ H_*(x,D)\neq y }  \Big ] = 1-\max_{\hat{y}\in\SY} p(y|x,D) = 1-\max_{\hat{y}\in\SY}\EE_{\theta\sim p(\theta|D)}\big[p(y|x,\theta )\big ] \:.
\]
The epistemic uncertainty:
\begin{eqnarray*}
  E_*(x,D) & = & \EE_{\theta \sim p(\theta|D)}\bigg [
  \EE_{y\sim p(y|x,\theta)} \Big [ 
  \one{ y\neq H_*(x,D)}
  - \one{ y\neq h_*(x,\theta)} 
  \Big ]
  \bigg ] \\
  &=& \EE_{y\sim p(y|x,D)}\Big[ \one{ y\neq H_*(x,D) } \Big ]
  - \EE_{\theta\sim p(\theta|D)}\bigg[ 
  \EE_{y\sim p(y|x,\theta)} \Big [ \one{ y \neq h_*(x,\theta) } \Big ]
  \bigg ] \\
  &=& \bigg (1- p\Big(y=H_*(x,D)\mid x,D\Big ) \bigg ) - \EE_{\theta\sim p(\theta|D)}\bigg [ 1- p\Big (y=h_*(x,\theta) \mid x,\theta\Big )\bigg ] \\
  & = & p\Big (y=h_*(x,\theta) \mid x,\theta\Big ) - p\Big(y=H_*(x,D)\mid x,D\Big ) \\
  & = & \EE_{\theta\sim p(\theta|D)}\bigg [\max_{y\in\SY} p(y\mid x,\theta) \bigg ] - \max_{y\in\SY} p(y|x,D) \:.
\end{eqnarray*}

%%%%%%%%%%%%%%%%%%%%%%%%%%%%%%%%%%
%%%%%%%%%%%%%%%%%%%%%%%%%%%%%%%%%%
\subsection{Cross-entropy loss}

Let $\SY=\{1,2,\ldots,Y\}$ be a finite set of labels.
Let the predictors $h\colon\SX\rightarrow \SP$ and $H\colon \SX\times(\SX\times\SY)^m\rightarrow\SP$ output a distribution over labels $\SY$, where $\SP=\{\#p\in\Re_+^Y\mid \sum_{y\in\SY}p_y=1\}$ denotes the probability simplex. Let $\ell(y,\#p)=-\log p_y$ denote the cross-entropy loss.

The Bayes-optimal predictor given the true parameter $\theta$ reads:
\[
   h_*(x,\theta) = \argmin_{\#p\in\SP} \EE_{y\sim p(y|x,\theta)} \big [-\log p_y \big ] = p(y|x,\theta) \:.
   %\argmin_{\#p\in\SP} \sum_{y\in\SY} p(y|x,\theta) (-\log p_y) = p(y|x,\theta) \:.
\]
The Bayesian predictor:
\[
   H_*(x,D) =\argmin_{\#p\in\SP} \EE_{y\sim p(y|x,D)} \big [-\log p_y \big ] = p(y|x,D)\:.
\]
The total uncertainty:
\[
  T_*(x,D) = \EE_{y\sim p(y\mid x,D)}\Big [-\log \big (p(y\mid x,D)\big) \Big] = \entropy{ p(y\mid x,D) }\:.
\]
The epistemic uncertainty:
\begin{eqnarray*}
  E_*(x,D) & = & \EE_{\theta \sim p(\theta|D)} \bigg [\EE_{y\sim p(y | x,\theta)} 
    \Big [
       -\log p(y\mid x,D) + \log p(y\mid x,\theta )
    \Big ]
  \bigg ]\\
  & = & \EE_{\theta \sim p(\theta|D)} \bigg [\EE_{y\sim p(y | x,\theta)} 
    \Big [ \log \Big ( \frac{p(y\mid x,\theta )}{p(y\mid x,D)} \Big )
    \Big ] \bigg ]\\
    & =& \EE_{\theta \sim p(\theta|D)} \bigg [
    {\rm D}_{\rm KL}\Big (p(y\mid x,\theta ) \,\|\,p(y\mid x,D) \Big )
    \bigg ]\:.
\end{eqnarray*}

\end{document}